\documentclass[10pt,twocolumn,letterpaper]{article}

\usepackage{cvpr}
\usepackage{times}
\usepackage{epsfig}
\usepackage{graphicx}
\usepackage{amsmath}
\usepackage{amssymb}

\usepackage[pagebackref=true,breaklinks=true,letterpaper=true,colorlinks,bookmarks=false]{hyperref}
\DeclareMathAlphabet{\pazocal}{OMS}{zplm}{m}{n}
\newcommand{\Lb}{\pazocal{L}}
\usepackage{enumitem}
\usepackage{cite}   
\usepackage{enumitem}
\usepackage{graphicx}
\usepackage{subfloat}
\usepackage{tabularx}
\usepackage{adjustbox}
\usepackage{amssymb}
\usepackage{multirow}
\usepackage[flushleft]{threeparttable}
\usepackage{subfigure}

\newcommand{\figref}[1]{Fig.~\ref{#1}}

\newcommand{\tabref}[1]{Table.~\ref{#1}}
\newcommand{\eqnref}[1]{Eq.~(\ref{#1})}
\newcommand{\secref}[1]{Sec.~\ref{#1}}
\renewcommand{\paragraph}[1]{\vspace{1mm}\noindent\textbf{#1}}

\renewcommand{\ie}{\textit{i.e.}}
\renewcommand{\eg}{\textit{e.g.}}

\cvprfinalcopy % *** Uncomment this line for the final submission

\def\cvprPaperID{1345} % *** Enter the CVPR Paper ID here

\ifcvprfinal\fi
\begin{document}

%%%%%%%%% TITLE
\title{Deep Video Inpainting}

\author{Dahun Kim\thanks{Both authors have contributed equally to this work.}\\
KAIST\\
\and
Sanghyun Woo\footnotemark[1]\\
KAIST\\
\and
Joon-Young Lee\\
Adobe Research\\
\and
In So Kweon\\
KAIST\\
}

\maketitle
%\thispagestyle{empty}

%%%%%%%%% ABSTRACT
\begin{abstract}
Video inpainting aims to fill spatio-temporal holes with plausible content in a video. Despite tremendous progress of deep neural networks for image inpainting, it is challenging to extend these methods to the video domain due to the additional time dimension. In this work, we propose a novel deep network architecture for fast video inpainting. Built upon an image-based encoder-decoder model, our framework is designed to collect and refine information from neighbor frames and synthesize still-unknown regions. At the same time, the output is enforced to be temporally consistent by a recurrent feedback and a temporal memory module. Compared with the state-of-the-art image inpainting algorithm, our method produces videos that are much more semantically correct and temporally smooth. In contrast to the prior video completion method which relies on time-consuming optimization, our method runs in near real-time while generating competitive video results. Finally, we applied our framework to video retargeting task, and obtain visually pleasing results.
\end{abstract}

%%%%%%%%% BODY TEXT
\section{Introduction}

Video inpainting can help numerous video editing and restoration tasks such as undesired object removal, scratch or damage restoration, and retargeting. More importatnly, and apart from its converntional demands, video inpainting can be used in combination with Augmented Reality (AR) for a greater visual experience; Removing existing items gives more opportunities before overlaying new elements in a scene. Therefore, as a Diminished Reality (DR) technology, it opens up new opportunities to be \textit{paired with recent real-time / deep learning-based AR technologies}. Moreover, there are several semi-online streaming scenarios such as automatic content filtering and visual privacy filtering. Only a small wait will lead to a considerable latency, thus making the speed itself an important issue.

\begin{figure}[t]
\def\arraystretch{0.5}
\includegraphics[width=1.0\linewidth]{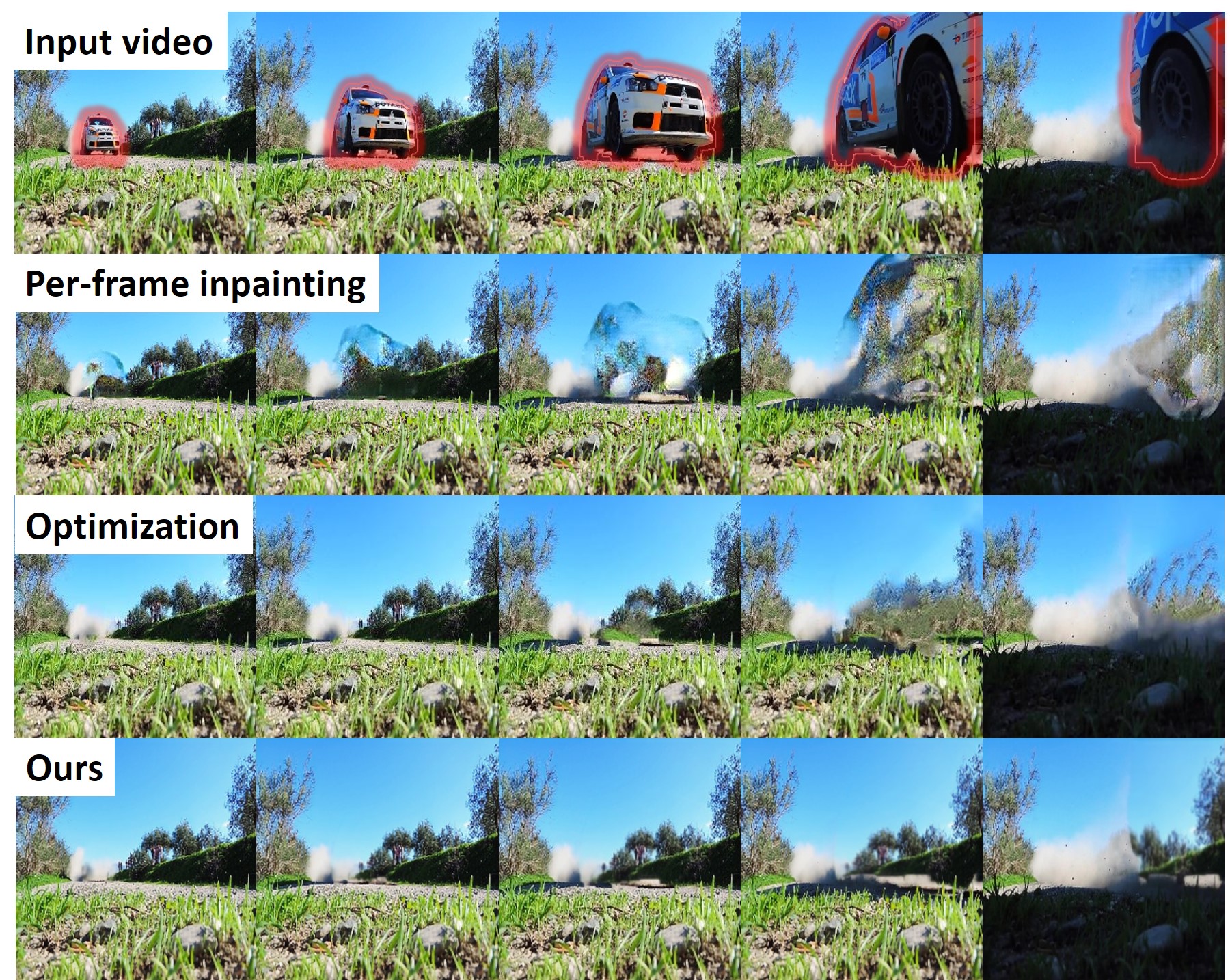}
\caption{Input video with mask boundaries in red (row-1). Video inpainting results by per-frame image inpainting~\cite{yu2018generative} (row-2), optimization-based method~\cite{huang2016temporally} (row-3), and our method (row-4). \textit{Best viewed when zoomed-in}.}
\label{fig:teaser}
\end{figure}

Despite tremendous progress on deep learning-based inpainting of a single image, it is still challenging to extend these methods to video domain due to the additional time dimension. The difficulties coming from complex motions and high requirement on temporal consistency make video inpainting a challenging problem. A straightforward way to perform video inpainting is to apply image inpainting on each frame individually. However, this ignores motion regularities coming from the video dynamics, and is thus incapable of estimating non-trivial appearance changes in image-space over time. Moreover, this scheme inevitably brings temporal inconsistencies and causes severe flickering artifacts. The second row in~\figref{fig:teaser} shows an example of directly applying the state-of-the-art feed-forward image inpainting~\cite{yu2018generative} in a frame-by-frame manner.

To address the temporal consistency, several methods have been developed to fill in the missing motion fields; using a greedy selection of local spatio-temporal patches~\cite{shiratori2006video}, a per-frame diffusion-based technique~\cite{matsushita2006full}, or an iterative optimization~\cite{huang2016temporally}. However, the first two methods treat flow estimation to be independent of color estimation~\cite{shiratori2006video,matsushita2006full} and the last relies on time-consuming optimization~\cite{huang2016temporally} (3rd row in \figref{fig:teaser}), which is effective but limits their practicality and flexibility in general scenarios.

% (video inpainting). + OPTIM. IGNORES SEMANTICS?

One might attempt to maintain temporal consistency by applying a post-processing method. Recently, Lai~\etal~\cite{lai2018learning} proposed a deep CNN model that takes both original and per-frame processed videos as input and produces a temporally consistent video. However, their method is only applicable when those two input videos have a pixel-wise correspondences (\eg colorization), which is not the case for video inpainting.

In this paper, we investigate whether a feed-forward deep network can be adapted to the video inpainting task. Specifically, we attempt to train a model with two core functions: 1) temporal feature aggregation and 2) temporal consistency preserving.
For the \textbf{temporal feature aggregation}, we cast the video inpainting task as a sequential multi-to-single frame inpainting problem. In particular, we introduce a novel 3D-2D feed-forward network which is built upon a 2D-based (image based) encoder-decoder model. The network is designed to collect and refine potential hints from neighbor frames and synthesize semantically-coherent video content in space and time. For the \textbf{temporal consistency}, we propose to use a recurrent feedback and a memory layer (\eg~convoutional LSTM~\cite{xingjian2015convolutional}). In addition, we use a flow loss to learn a warping of the previously synthesized frame and a warping loss to enforce both short-term and long-term consistency in results. Finally, we come up with a single, unified deep CNN model called \textbf{VINet}. 

We conduct extensive experiments to validate the contributions of our design choices. We show that our multi-to-single frame formulation produces videos that are much more accurate and visually pleasing than the method of~\cite{yu2018generative}. An example result of our method is shown in the last row of~\figref{fig:teaser}. Our model sequentially processes video frames of arbitrary length and requires no optical flow computation at the test time, thus runs at a near real-time rate.

\paragraph{Contribution.} In summary, our contribution is as follow.
\begin{enumerate}[topsep=0pt,itemsep=0pt]
\item  We cast video inpainting as a sequential multi-to-single frame inpainting task and present a novel deep 3D-2D encoder-decoder network. Our method effectively gathers features from neighbor frames and synthesizes missing content based on them.
\item  We use a recurrent feedback and a memory layer for the temporal stability. Along with the effective network design, we enforce strong temporal consistency via two losses: flow loss and warping loss.
\item  Up to our knowledge, it is the first work to provide a single, unified deep network for the general video inpainting task. We conduct extensive subjective and objective evaluations and show its efficacy. Moreover, we apply our method to video retargeting and super-resolution tasks, demonstrating favorable results.
\end{enumerate}
%-------------------------------------------------------------------

\section{Related Work}
\label{sec:related}

Significant progress has been made on image inpainting~\cite{ballester2001filling,bertalmio2000image,efros1999texture,pathak2016context,yang2017high,yeh2017semantic,iizuka2017globally,yu2018generative,liu2018image,yu2018free}, to a point of where commercial solutions are now available~\cite{barnes2009patchmatch}. However, video inpainting algorithms have been under-investigated. This is due to the additional time dimension which introduces major challenges such as severe viewpoint changes, temporal consistency preserving, and high computational complexity. Most recent methods found in the literature address these issues using either object-based or patch-based approaches.

In object-based methods, a pre-processing is required to split a video into foreground objects and background, and it is followed by an independent reconstruction and merging step at the end of algorithms. Previous efforts which fall under this category are homography-based algorithms that are based on the graph-cut~\cite{granados2012not,granados2012background}. However, the major limitation of these object-based methods is that the synthesized content has to be copied from the visible regions. Therefore, these methods are mostly vulnerable to abrupt appearance changes such as scale variations, \eg~when an object moves away from the camera.

In patch-based methods, the patches from known regions are used to fill in a mask region. For example, Patwardhan~\etal~\cite{patwardhan2005video,patwardhan2007video} extend the well-known texture synthesis technique~\cite{efros1999texture} to video inpainting. However, these methods either assume static cameras~\cite{patwardhan2005video} or constrained camera motion~\cite{patwardhan2007video} and are based on a greedy patch-filling process where the early errors are inevitably propagated, yielding globally inconsistent outputs.

To ensure the global consistency, patch-based algorithms have been cast as a global optimization problem. Wexler~\etal~\cite{wexler2004space} present a method that optimizes a global energy minimization problem for 3D spatio-temporal patches by alternating between patch search and reconstruction steps. Newson~\etal~\cite{newson2014video} extend this by developing a spatio-temporal version of PatchMatch~\cite{barnes2009patchmatch} to strengthen the temporal coherence and speed up the patch matching. Recently, Huang~\etal~\cite{huang2016temporally} modify the energy term of ~\cite{wexler2004space} by adding an optical flow term to enforce temporal consistency. Although these methods are effective, their biggest limitations are high computational complexity and the absolute dependence upon the pre-computed optical flow which cannot be guaranteed to be accurate in complex sequences.

To tackle these issues, we propose a deep learning based method for video inpainting. To better exploit temporal information coming from multiple frames and be highly efficient, we construct a 3D-2D encoder-decoder model, that can provide traceable features revealed from the video dynamics. It takes total 6 frames as input; 5 source frames and 1 reference frame (\ie the frame to be inpainted). We learn the feature flow between frames to deal with both hole-filling and coherence. The still-unknown regions are synthesized in a semantically natural way based on the surrounding context. We argue that our method provides a better prospect than the previous optimization-based techniques in that deep CNNs are excellent at learning spatial semantics and temporal dynamics from an ever-growing vast amount of video data. To our best knowledge, this is the first work that deeply addresses the general video inpainting problem via a deep CNN model.

\begin{figure*}[t]
\begin{tabular}{@{}c@{}}
\includegraphics[width=0.96\linewidth]{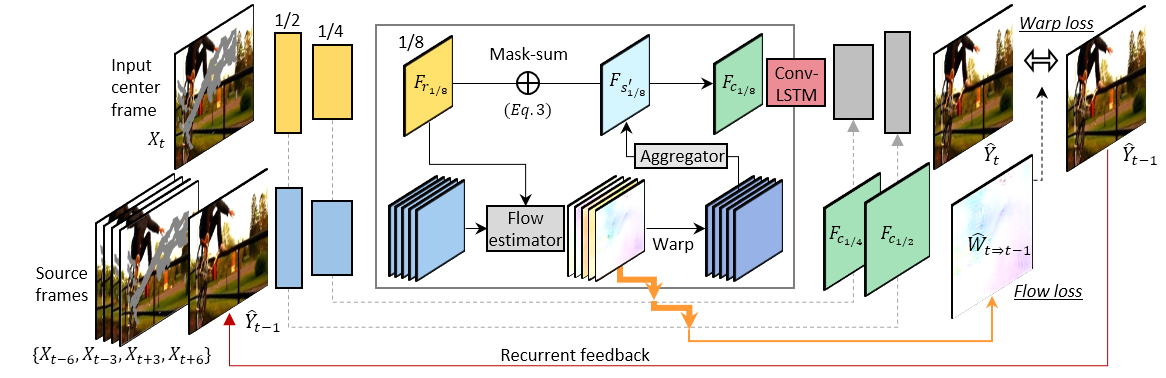} \\
\end{tabular}
\caption{ {\bf The overview of VINet.} Our network takes in multiple frames (${X}_{t-6}, {X}_{t-3}$, ${X}_{t}$, ${X}_{t+3}, {X}_{t+6}$) and the previously generated frame ($\hat{Y}_{t-1}$), and generates the inpainted frame ($\hat{Y}_{t}$) as well as the flow map ($\hat{W}_{t\Rightarrow t-1}$). We employ both flow sub-networks and mask sub-networks at 4 scales (1/8, 1/4, 1/2, and 1) to aggregate and synthesize feature points progressively. For temporal consistency, we use a recurrent feedback and a temporal memory layer (ConvLSTM) along with two losses: flow loss and warp loss. The orange arrows denote the $\times 2$ upsampling for residual flow learning as in~\cite{sun2018pwc} for 5 streams, while the thinner orange arrow is for only the stream from $\hat{Y}_{t-1}$. The mask sub-networks are omitted in the figure for the simplicity.}
\label{fig:architecture}
\end{figure*}

\section{Method}
\label{sec:method}
% In this section, we formulate the problem of video inpainting, present our network design, and describe our two-stage training scheme with the used loss functions.

\subsection{Problem Formulation}
Video inpainting aims to fill in arbitrary missing regions in video frames $X_1^T := \{X_1, X_2, ..., X_T\}$. The reconstructed regions should be either accurate as in the ground-truth frames $Y_1^T := \{Y_1, Y_2, ..., Y_T\}$ and consistent in space and time. We formulate the video inpainting problem as learning a mapping function from $X_1^T$ to the output $\hat{Y}_1^T := \{ \hat{Y}_1, \hat{Y}_2, ...,\hat{Y}_T\}$ such that the conditional distribution $p(\hat{Y}_1^T|X_1^T)$ is identical to $p(Y_1^T|X_1^T)$. Through matching the conditional distributions, the network learns to generate realistic and temporally-consistent output sequences. To simplify the problem, we make a Markov assumption where we factorize the conditional distribution to a product form. In this form, the naive \textit{frame-by-frame} inpainting can be formulated as
\begin{equation}
    p(\hat{Y_1^T}|X_1^T) = \prod_{t=1}^{T} p(\hat{Y}_t|X_t).
\end{equation}

However, to obtain visually pleasing video results, we argue that the generation of $t$-th frame $\hat{Y_t}$ should be consistent with 1) spatio-temporal neighbor frames $X_{t-N}^{t+N}$ where $N$ denotes a temporal radius, 2) the previously generated frame $\hat{Y}_{t-1}$ and 3) all previous history encoded in a recurrent memory $M_t$. Thus, we propose to learn the conditional distribution of
\begin{equation}
    p(\hat{Y_1^T}|X_1^T) = \prod_{t=1}^{T} p(\hat{Y}_t|X_{t-N}^{t+N}, \hat{Y}_{t-1}, M_t).
\label{eqn:our_formulation}
\end{equation}

In our experiments, we set $N$ to 2, taking two lagging and two leading frames to recover the current frame. We sample frames with a temporal stride 3, such that $X_{t-N}^{t+N} := \{ {X}_{t-6}, {X}_{t-3}, {X}_{t}, {X}_{t+3}, {X}_{t+6}\}$. We want to recover the current frame by both aggregating information from neighbor frames and synthesizing totally blind regions jointly.
%Our strategy is not only to recover from the current frame, but also collect and refine potential hints from neighbor frames ($X_{t-N}^{t+N}$) and synthesize still-unknown regions. 
At the same time, the output is enforced to be temporally consistent with the past predictions by the recurrent feedback ($\hat{Y}_{t-1}$) and the memory ($M_t$). We train a deep network $D$ to model the conditional distribution $p(\hat{Y}_t|X_{t-N}^{t+N}, \hat{Y}_{t-1}, M_t)$ as $\hat{Y}_t=D(X_{t-N}^{t+N}, \hat{Y}_{t-1}, M_t)$. We obtain the final output $\hat{Y}_1^T$ by applying the function $D$ in an autoregressive manner. Our multi-to-single frame formulation outperforms a single-frame baseline and even produces results comparable with the optimization-based method, as described in \secref{sec:experiments}.

\subsection{Network Design}

Our full model (VINet) jointly learns to inpaint the video and maintain temporal consistency. The overview of VINet is illustrated in ~\figref{fig:architecture}.

\subsubsection{Multi-to-Single Frame Video Inpainting}
In videos, the occluded or removed parts in a frame are often revealed in the past/future frames as the objects move and the viewpoint changes. If such hints exist in the temporal radius, those disclosed content can be borrowed to recover the current frame. Otherwise, the still-unknown regions should be synthesized. To achieve this, we construct our model as an encoder-decoder network that learns such temporal feature aggregation and single-frame inpainting simultaneously. The network is designed to be fully convolutional, which can handle arbitrary size input.

\paragraph{Source and reference encoders.} 
The encoder is a multiple-tower network with source and reference streams. The source stream takes past and future frames with the inpainting masks as input. For the reference stream, the current frame and its inpainting mask are provided. We concatenate the image frames and the masks along the channel axis, and feed into the encoder.
% \figref{fig:2_tower} shows an (toy) example of 2-tower encoder with one source stream. 
In practice, we use a 6-tower encoder: 5 source streams with weight-sharing that take two lagging (${X}_{t-6}, {X}_{t-3}$) and two leading frames (${X}_{t+3}, {X}_{t+6}$), and the previously generated frame ($\hat{Y}_{t-1}$), and 1 reference stream. The source features that are non-overlapping with the reference features can be borrowed to inpaint the missing regions by the following feature flow learning and learnable feature composition.

\paragraph{Feature flow learning.}
Before directly combining the source and reference features, we propose to explicitly align the feature points. This strategy helps our model easily borrow traceable features from the neighbor frames. To achieve this, we insert flow sub-networks to estimate the flows between the source and reference feature maps in four different spatial scales (1/8, 1/4, 1/2, and 1). We adopt the coarse-to-fine structure of PWCNet~\cite{sun2018pwc}. The explicit flow supervision is only given at the finest scale (\ie~1) and \textit{only between} the consecutive two frames, where we extract the pseudo-ground-truth flow ${W}_{t\Rightarrow t-1}$ between ${Y}_{t}$ and ${Y}_{t-1}$  using FlowNet2~\cite{ilg2017flownet}.

\paragraph{Learnable Feature Composition.} 
Given the aligned feature maps from the five source streams, they are concatenated along the time dimension and fed into a $5\times3\times3$ (THW) convolution layer that produces a spatio-temporally aggregated feature map $F_{s'}$ with the time dimension of 1. This is designed to dynamically select source feature points across the time axis, by highlighting the features complementary to the reference features and ignoring otherwise. For each 4 scales, we employ a mask sub-network to combine the aggregated feature map $F_{s'}$ with the reference feature map $F_{r}$. The mask sub-network consists of three convolution layers and takes the absolute difference of the two feature maps $|F_{s'} - F_{r}|$ as input and produces single channel composition mask $m$, as suggested in~\cite{chen2017coherent}. By using the mask, we can gradually combine the warped features and the reference features. At the scale of 1/8, the composition is done by 
\begin{equation}
    F_{c_{1/8}} = (1-m_{1/8}) \odot F_{r_{1/8}} + m_{1/8} \odot F_{s'_{1/8}},
\label{eqn:feature_composition}
\end{equation}
where $\odot$ is the element-wise product operator. 

\paragraph{Decoder.} 
To pass image details to the decoder, we employ skip connections as in U-net~\cite{ronneberger2015u}. To prevent the concern raised by~\cite{yu2018free} that skip connections contain zero values at the masked region, our skip-connections pass the composite features similarly to \eqnref{eqn:feature_composition}, as %our skip-connecting features on these regions are composed of features warped from the source streams similar to \eqnref{eqn:feature_composition}
\begin{eqnarray}
    F_{c_{1/4}} = (1-m_{1/4}) \odot F_{r_{1/4}} + m_{1/4} \odot F_{s'_{1/4}},\\
    F_{c_{1/2}} = (1-m_{1/2}) \odot F_{r_{1/2}} + m_{1/2} \odot F_{s'_{1/2}}.
\end{eqnarray}

At the finest scale, the estimated optical flow $\hat{W}_{t\Rightarrow t-1}$ is used to warp the previous output $\hat{Y}_{t-1}$ to the current raw output $\hat{Y'}_{t}$. We then blend this warped image and the raw output with the composition mask $m_{1}$, to obtain our final output $\hat{Y}_{t}$ as
\begin{equation}
    \hat{Y}_{t} = (1-m_{1}) \odot \hat{Y'}_{t} + m_{1} \odot \hat{W}_{t\Rightarrow t-1}(\hat{Y}_{t-1}).
\end{equation}

\subsubsection{Recurrence and Memory}
To strongly enforce the temporal coherence on the video output, we propose to use the recurrent feedback loop ($\hat{Y}_{t-1}$) and the temporal memory layer ($M_{t}$) as formulated in \eqnref{eqn:our_formulation}. 

Our formulation encourages the current output to be conditional to the previous output frame. The knowledge from the previous output encourages the traceable features to be kept unchanged, while the untraceable (\eg~occlusion) points to be synthesized. This not only helps the output to be consistent along the motion trajectories but also avoids ghosting artifacts at occlusions or motion discontinuities.

While the recurrent feedback connects the consecutive frames, filling in the large holes requires more long-term (\eg~5 frames) knowledge. 
At this point, the temporal memory layer can help to connect internal features from different time steps in the long term. We adopt a convolutional LSTM (ConvLSTM) layer and a warping loss as suggested in~\cite{lai2018learning}. In particular, we feed the composite feature $F_{c}$ at the scale 1/8 to the ConvLSTM at every time step.

\subsection{Losses}
We train our network to minimize the following loss function,
%{\small
\begin{eqnarray}
     \mathcal{L} = \lambda_{R}\mathcal{L}_{R}+\lambda_{F}\mathcal{L}_{F}+\lambda_{W}\mathcal{L}_{W},
\label{eqn:total_loss}
\end{eqnarray}
%}
where $\mathcal{L}_{R}$ is the reconstruction loss, $\mathcal{L}_{F}$ is the flow estimation loss, and $\mathcal{L}_{W}$ is the warping loss. The balancing weights $\lambda_{R}, \lambda_{F}, \lambda_{W}$ are set to 1, 10, 1 respectively throughout the experiments. For the temporal losses   $\mathcal{L}_{F}$ and $\mathcal{L}_{W}$, we set the number of recurrences as 5 $(T = 5)$.
% Please refer to supplementary material for the detailed specifications of each loss term.

$\mathcal{L}_{R}$ consists of two terms, $\mathcal{L}_{1}$ and $\mathcal{L}_{ssim}$,
%{\footnotesize
\begin{eqnarray}
\Lb_{1} = \left \|  \hat{Y_t} - Y_t \right\|_{1}, \\ 
\Lb_{ssim} =  (\frac{{(2\mu_{\hat{Y_t}} \mu_{Y_t} + c_1 )(2\sigma_{\hat{Y_t} {Y_t}} + c_2)}}
{{(\mu_{\hat{Y_t}}^2+\mu_{Y_t}^2 +c_1)(\sigma_{\hat{Y_t}}^2+\sigma_{Y_t}^2 +c_2)}}), \\
%\Lb_{grad} = \left \|\nabla_{W}(\hat{Y_t} - Y_t)\right\|_{1} + \left \|\nabla_{H}(\hat{Y_t} - Y_t)\right\|_{1}, \\
\Lb_{R} = \Lb_{1} + \Lb_{ssim},
\end{eqnarray}
%}
where $\hat{Y_t}, Y_t$ denote the predicted frame and the ground-truth frame respectively. $\mu, \sigma$ denote the average, variance, respectively. $c_1, c_2$ denote the stabilization constants which are respectively set to $0.01^2, 0.03^2$.

The flow loss $\mathcal{L}_{F}$ is defined as
%{\footnotesize
\begin{equation}
\sum\limits_{t=2}^{T} (\left \| {W}_{t\Rightarrow t-1} - \hat{W}_{t\Rightarrow t-1} \right\|_{1} + \left \| {Y}_{t} - \hat{W}_{t\Rightarrow t-1}({Y}_{t-1}) \right\|_{1}),
\label{eqn:flowloss}
\end{equation}
%}
where ${W}_{t\Rightarrow t-1}$ is the pseudo-ground-truth backward flow between the target frames, ${Y}_{t}$ and ${Y}_{t-1}$, extracted by FlowNet2~\cite{ilg2017flownet}. In \eqnref{eqn:flowloss}, the first term is the endpoint error between the groundturth and the estimated flow, and the second is the warping error when the flow is used to warp the previous target frame to the next target frame.

The warping loss $\mathcal{L}_{W}$ includes $\mathcal{L}_{st}$ and $\mathcal{L}_{lt}$ as,
%{\footnotesize
\begin{eqnarray}
\Lb_{st} = \sum\limits_{t=2}^T M_{t \Rightarrow t-1} \left \| \hat{Y_t} - {W}_{t \Rightarrow t-1}({Y}_{t-1})\right\|_{1}, \\
\Lb_{lt} = \sum\limits_{t=2}^T M_{t \Rightarrow 1} \left \| \hat{Y_t} - {W}_{t \Rightarrow 1}({Y}_{1})\right\|_{1}, \\
\Lb_{W} = \Lb_{st} + \Lb_{lt}.
\end{eqnarray}
%}
We follow the protocol in~\cite{lai2018learning} that uses FlowNet2~\cite{ilg2017flownet} to obtain $M_{t \Rightarrow t-1}$ and $W_{t-1}$, which respectively denote the binary occlusion mask and the backward optical flow between the target frames ${Y}_{t}$ and ${Y}_{t-1}$. We adopt both short-term and long-term temporal losses. Note that we use ground-truth target frames in the warping operation since the synthesizing ability is imperfect during training.

\subsection{Two-Stage Training}

We employ a two-stage training scheme that gradually learns the core functionalities for video inpainting; 1) We first train the model without the recurrent feedback and memory to focus on learning the temporal feature aggregation. At this stage, we only use the reconstruction loss $\mathcal{L}_{R}$; 2) We then add the recurrent feedback and the ConvLSTM layer, and fine-tune the model using the full loss function (\eqnref{eqn:total_loss}) for temporally coherent predictions. We use videos in the Youtube-VOS dataset~\cite{xu2018youtube} as ground-truth for the training. It is a large-scale dataset for video object segmentation containing 4000+ YouTube videos with 70+ common objects. All video frames are resized to $256\times256$ pixels for training and testing. %We create the inpainting masks as follows.

\paragraph{Video mask dataset.} 
In general video inpainting, the spatio-temporal holes consist in diverse motion and shape changes. To simulate this complexity during training, we create the following four types of video masks.

\begin{enumerate}[topsep=1pt,itemsep=1pt]
\item {Random square}: 
We randomly mask a square box in each frame. The visible regions each of input frames are mostly complementary so that the network can clearly learn how to align, copy, and paste neighboring feature points.
\item {Flying square}: 
The motion of the inpainting holes is rather regularized than random in real scenarios. To simulate such regularity, we shift a square by a uniform step size in one direction across the input frames. 
\item {Arbitrary mask}: To simulate diverse hole shapes and sizes, we use the irregular mask dataset~\cite{liu2018image} which consists of random streaks and holes of arbitrary shapes. During training, we apply random transformations (translation, rotation, scaling, sheering).
\item {Video object mask}: In the context of the video object removal task, masks with the most realistic appearance and motion can be obtained from video object segmentation datasets. We use the foreground segmentation masks of the YouTube-VOS dataset~\cite{xu2018youtube}.  
\end{enumerate}

\subsection{Inference}
We assume that the inpainting masks for all video frames are given. To avoid any data overlap between training and testing, we obtain object masks from the DAVIS dataset~\cite{perazzi2016benchmark, pont20172017}, the public benchmark dataset for video object segmentation. It contains dynamic scenes, complex camera movements, motion blur effects, and large occlusions. The inpainting mask is constructed by dilating the ground-truth segmentation mask. Our method processes frames recursively in a sliding window manner. 

% and runs at 15 fps on a GPU for frames of resolution $256 \times 256$.

\begin{figure}[t]
\def\arraystretch{1.0}
\begin{tabular}{c@{}}
    \includegraphics[width=1.0\columnwidth]{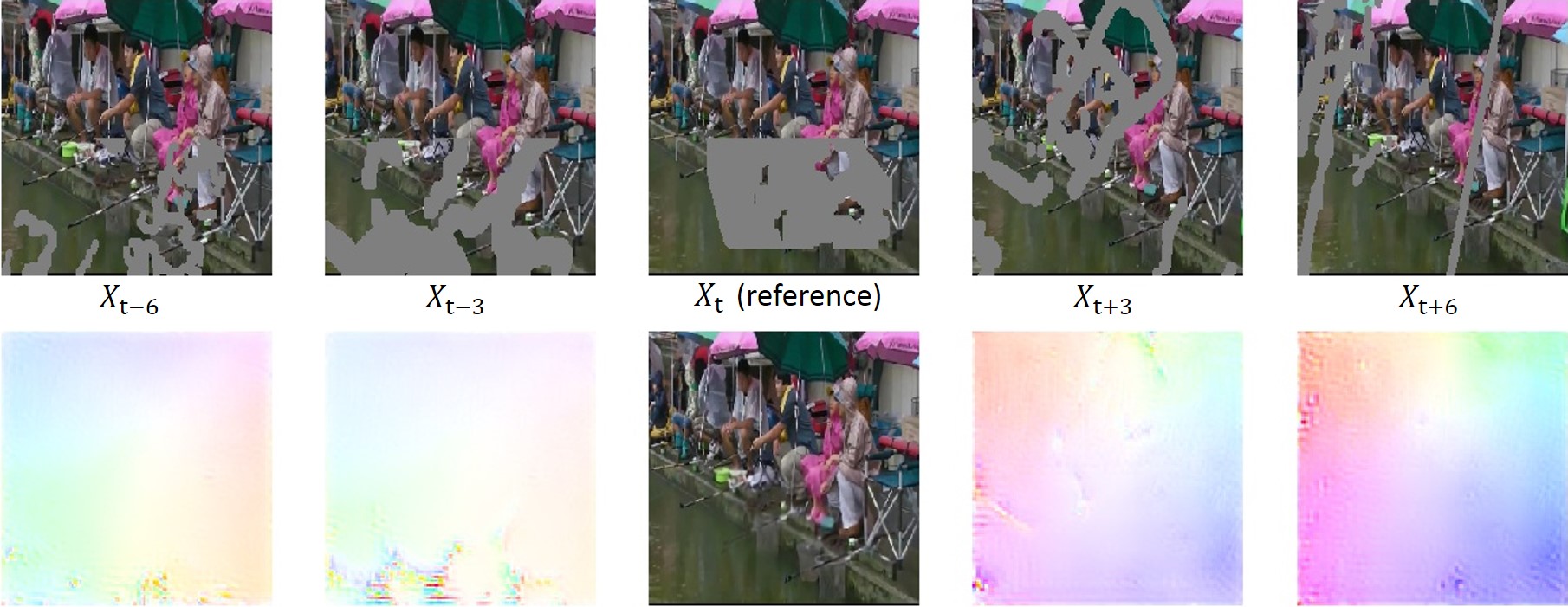} \\
    (a) \\
    \includegraphics[width=1.0\columnwidth]{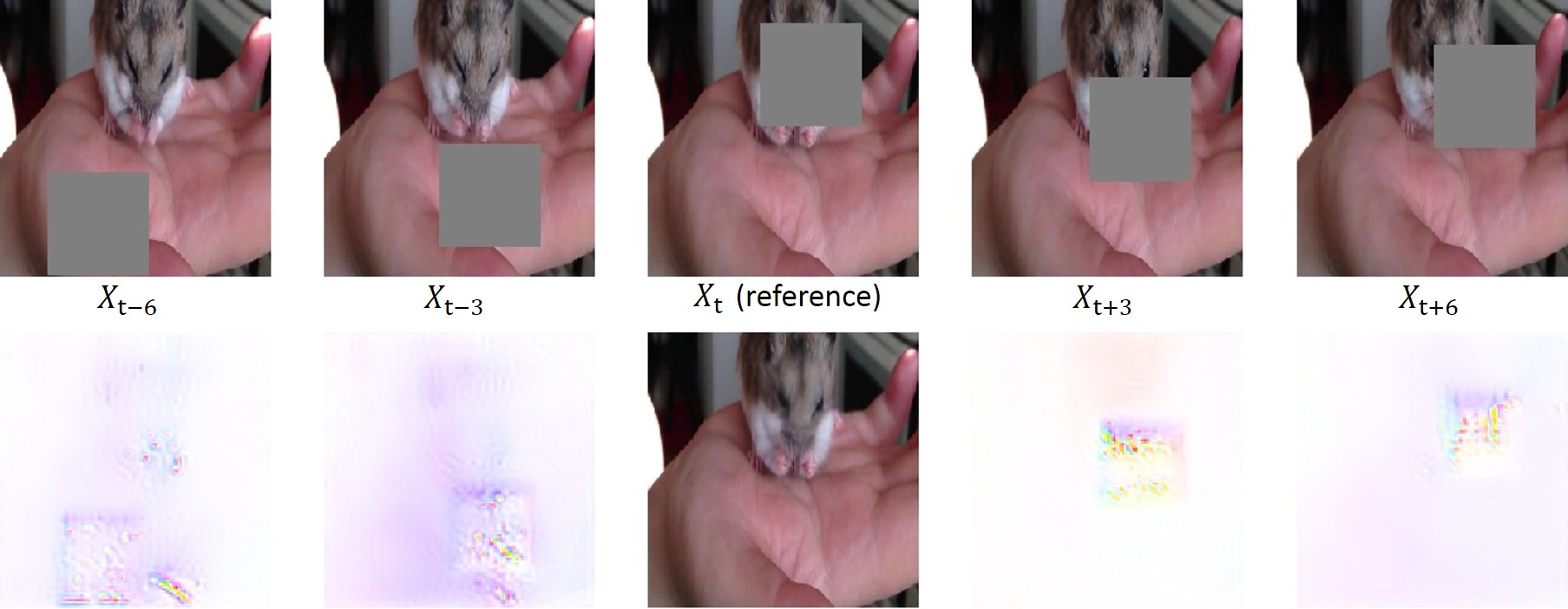}\\
    (b) \\
\end{tabular}
\caption{\textbf{Visualization of the learned feature composition.} Input frames are on the odd rows, and corresponding feature flows referential to the center, and the inpainted frame are on the even rows. Our network successfully aligns and integrates the source features to fill in the large and complex hole in the reference frame.}
\label{fig:vis_feature_comp}
\end{figure}

\subsection{Implementation Details}
Our model is implemented using Pytorch v0.4, CUDNN v7.0, CUDA v9.0. It run on the hardware with Intel(R) Xeon(R) (2.10GHz) CPU and NVIDIA GTX 1080 Ti GPU. The model runs at 12.5 fps on a GPU for frames of $256\times256$ pixels. We use Adam optimizer with $\beta$ = (0.9, 0.999) and a fixed learning rate 1e-4. We train our model from scatch. The first and second training stage takes about 1 day each using four NVIDIA GTX 1080 Ti GPUs.

% We skipped frames with a stride to maximize the gain from the scene dynamics.
% In our preliminary experiment, we found the sampling interval of 3 performs the best. In the second training stage, we set the number of recurrences as 5. 

\section{Experiments}
\label{sec:experiments}
In this section, we conduct experiments to analyze our two major design choices. Specifically, we visualize the learned multi-to-single mechanism and show the impact of the added recurrence and memory. We then evaluate our video results both quantitatively and qualitatively, compared with the state-of-the-art baselines. Finally, we demonstrate the applicability of our framework on video retargeting and video super-resolution tasks. 

\paragraph{Baselines.} We compare our approach to two state-of-the-art baselines in the literature by running their test codes with our testing videos and masks.
\begin{itemize}[topsep=1pt,itemsep=1pt]
\item Yu~\etal~\cite{yu2018generative}: A feed-forward CNN based method, which is designed for single image inpainting. We processes videos frame-by-frame without using any temporal information.
\item Huang~\etal~\cite{huang2016temporally}: An optimization-based video completion method, which jointly estimates global flow and color. It requires on-the-fly optical flow computation and is extremely time-consuming.
\end{itemize}

\subsection{Visualization of Learned Feature Composition}
% 1ST DESIGN CHOICE MULTI-TO-SINGLE

\figref{fig:vis_feature_comp} shows that the proposed model explicitly borrows visible neighbor features to synthesize the missing content. 
For the visualization, we take the model of the first training stage and plot the learned feature flow from each of the four source streams to the reference stream, at $128\times128$ pixel resolution.
We observe that even with a large and complex hole in the reference (center) frame, our network is able to align the source feature maps to the reference and integrate them to fill in the hole. Even without an explicit flow supervision, our flow sub-network is able to warp the feature points in visible regions and shrink the unhelpful zero features in masked regions. Moreover, these potential hints are adjusted according to the spatio-temporal semantics, rather than copied-and-pasted in a fixed manner. One example is shown in \figref{fig:vis_feature_comp}-(b) where the eyes of the hamster are synthesized \textit{half-closed}.

\subsection{Improvement on Temporal Consistency}
\label{sec:TC}
% 2ND DESIGN CHOICE, RECURRENCE and MEMORY
We compare the temporal consistencies of our video results before and after adding the recurrent feedback and the convLSTM. To validate the effectiveness of our method, we also compare with the two representative baselines mentioned above~\cite{yu2018generative,huang2016temporally}. Since the Sintel dataset~\cite{butler2012naturalistic} provides ground-truth optical flows, we use it to quantitatively measure the \textit{flow warping errors}\cite{lai2018learning}.  We use the object masks in the DAVIS dataset~\cite{perazzi2016benchmark, pont20172017} as our inpainting mask sequences. %The average warping error of a video sequence is used for the evaluation.
We take 32 frames each from 21 videos in Sintel to constitute our inputs and experiment for five trials. For each trial, we randomly select 21 videos of length 32+ from DAVIS to create corresponding mask sequences and keep them unchanged for all the methods. 

% {\small
% \begin{eqnarray}
% \frac{1}{T-1}\sum\limits_{t=1}^{T-1} (\left \| {Y}_{t+1} - \hat{W}_{t+1\Rightarrow t}({Y}_{t}) \right\|_{1})
% \end{eqnarray}
% }

In \tabref{tab:warping}, we report the flow warping errors averaged over the videos and trials. It shows that our full model outperforms other baselines by large margins. Even the global (heavy) optimization method~\cite{huang2016temporally} performs marginally better than our 1st-stage method and has a much larger error than our full model. Not surprisingly, Yu~\etal's method turns out to be the least temporally consistent. Note that the error of our full model is reduced by a factor of 10 after adding the recurrent feedback and the convLSTM layer, implying that they significantly improve the temporal stability in the short and long term.

\begin{table}
\centering
%\resizebox{\linewidth}{!}{%
\setlength{\tabcolsep}{10pt}{
\normalsize
\begin{tabular}{ l|c }
\hline
& DAVIS masks on Sintel frames \\
\hline
Frame-by-frame~\cite{yu2018generative}            & 0.0429\\
Optimization~\cite{huang2016temporally}  & 0.0343\\
\textbf{VINet} (agg. only)                                  & 0.0383 \\
\textbf{VINet} (agg. + T.C.)              & \textbf{0.0015}  \\
\hline
\end{tabular}
}
\caption{\textbf{Flow warping errors.} We evaluate the flow warping errors on the Sintel dataset using 21 videos and ground truth flows.}
\label{tab:warping}
\end{table}

\begin{table}
\centering
%\resizebox{\linewidth}{!}{%
\setlength{\tabcolsep}{10pt}{
\normalsize
\begin{tabular}{ l|c }
\hline
& DAVIS masks on DAVIS frames \\
\hline
Frame-by-frame~\cite{yu2018generative} & 0.0080  \\
Optimization~\cite{huang2016temporally} & 0.0053\\
% Multi-to-Single& 0.0073 \\
\textbf{VINet} (agg. only) & 0.0073 \\
\textbf{VINet} (agg. + T.C.) & \textbf{0.0046} \\
\hline
\end{tabular}
}
\caption{\textbf{FID scores.} We evaluate the FID scores on the DAVIS dataset using 20 videos.}
\label{tab:FID}
\end{table}

\begin{figure*}[t]
\begin{center}
\def\arraystretch{1.0}
\begin{tabular}{@{}c@{\hskip 0.01\linewidth}c@{}}
    \includegraphics[width=0.495\linewidth]{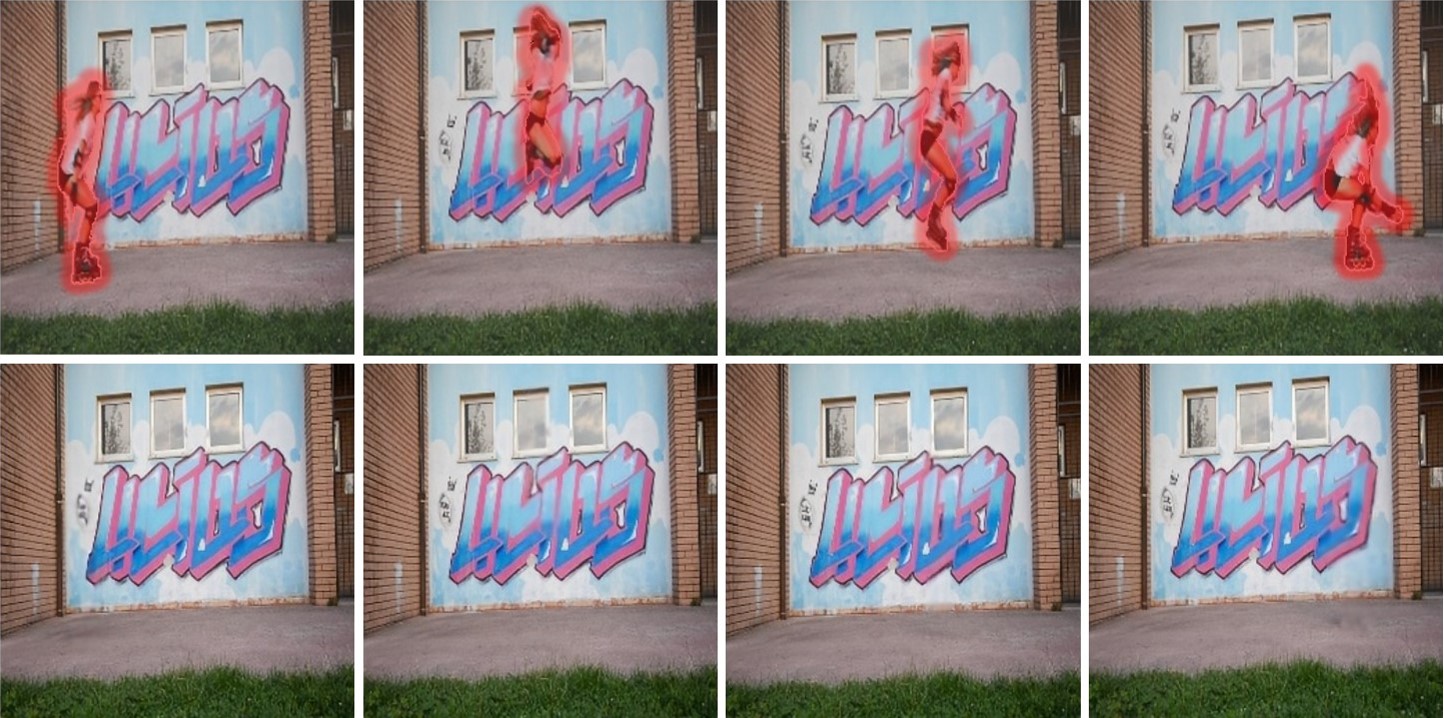}&
    \includegraphics[width=0.495\linewidth]{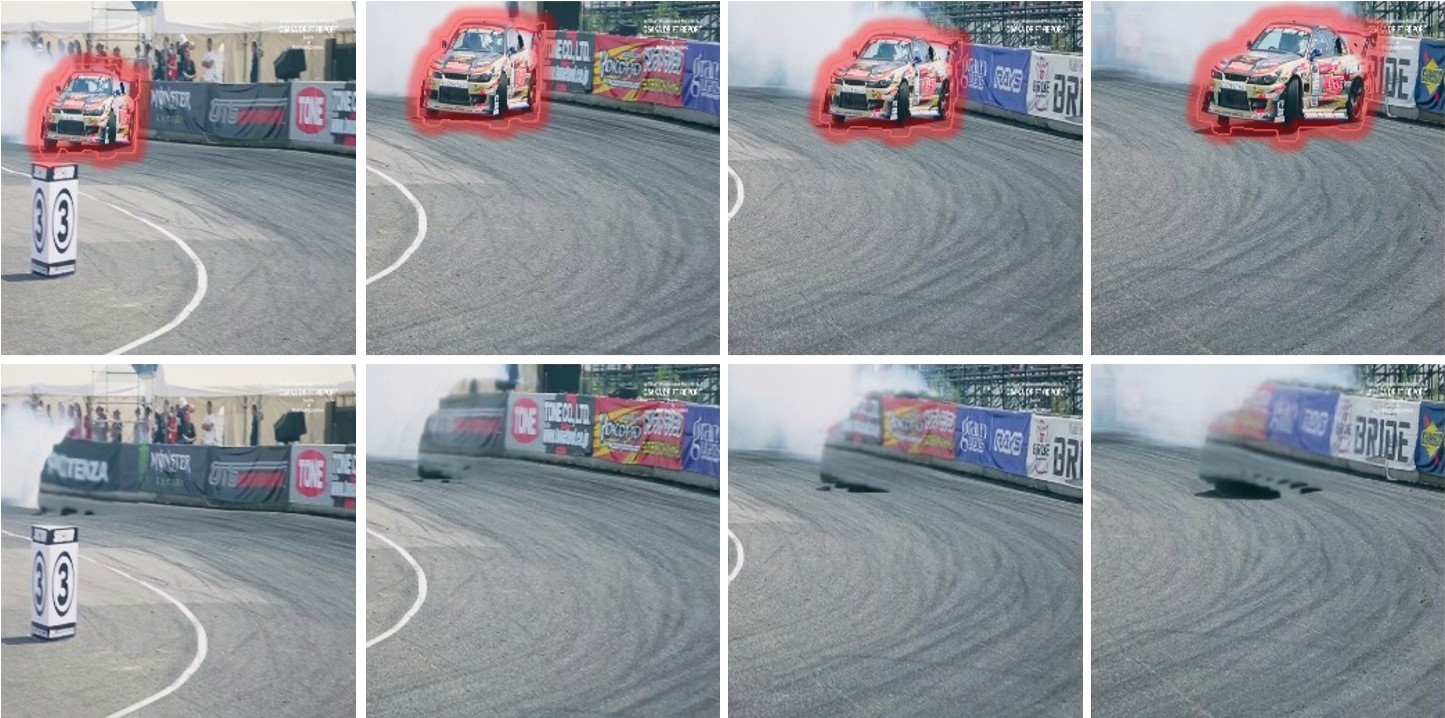}\\
    \includegraphics[width=0.495\linewidth]{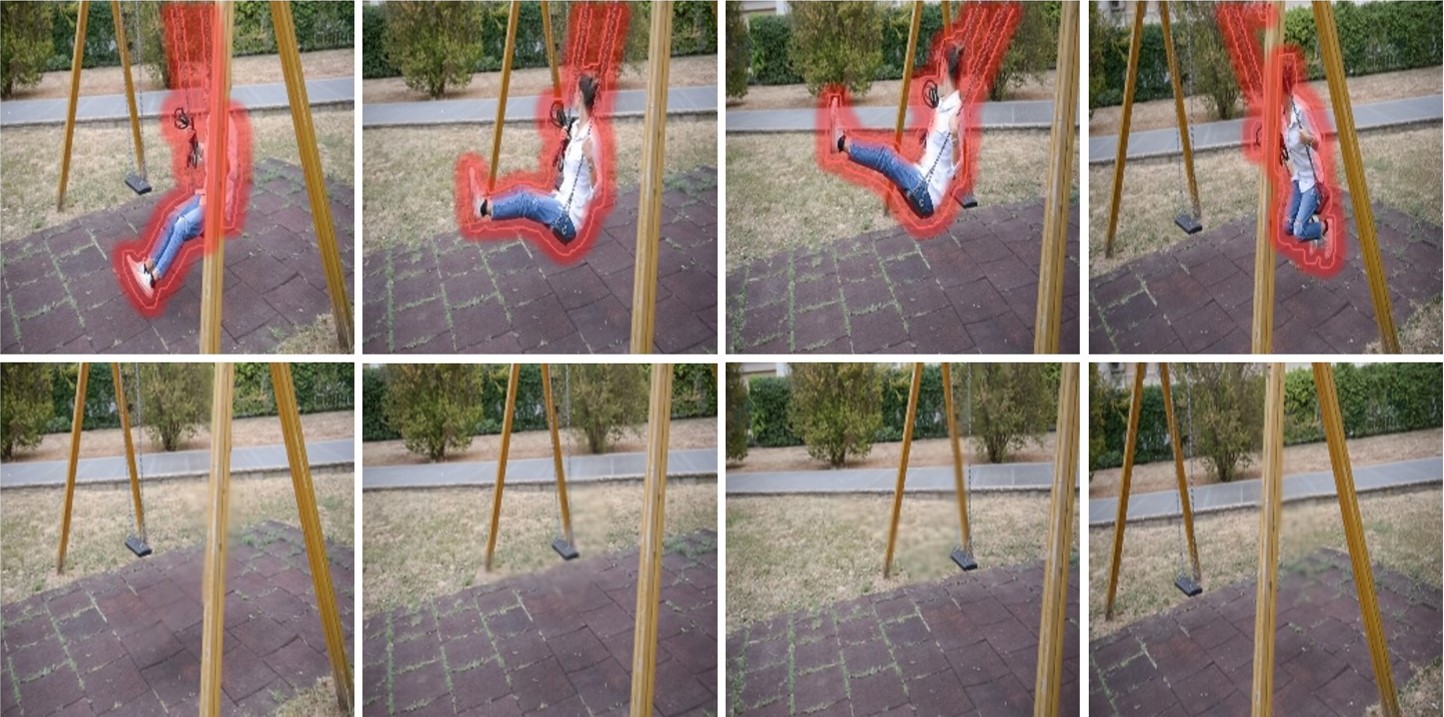} & 		\includegraphics[width=0.495\linewidth]{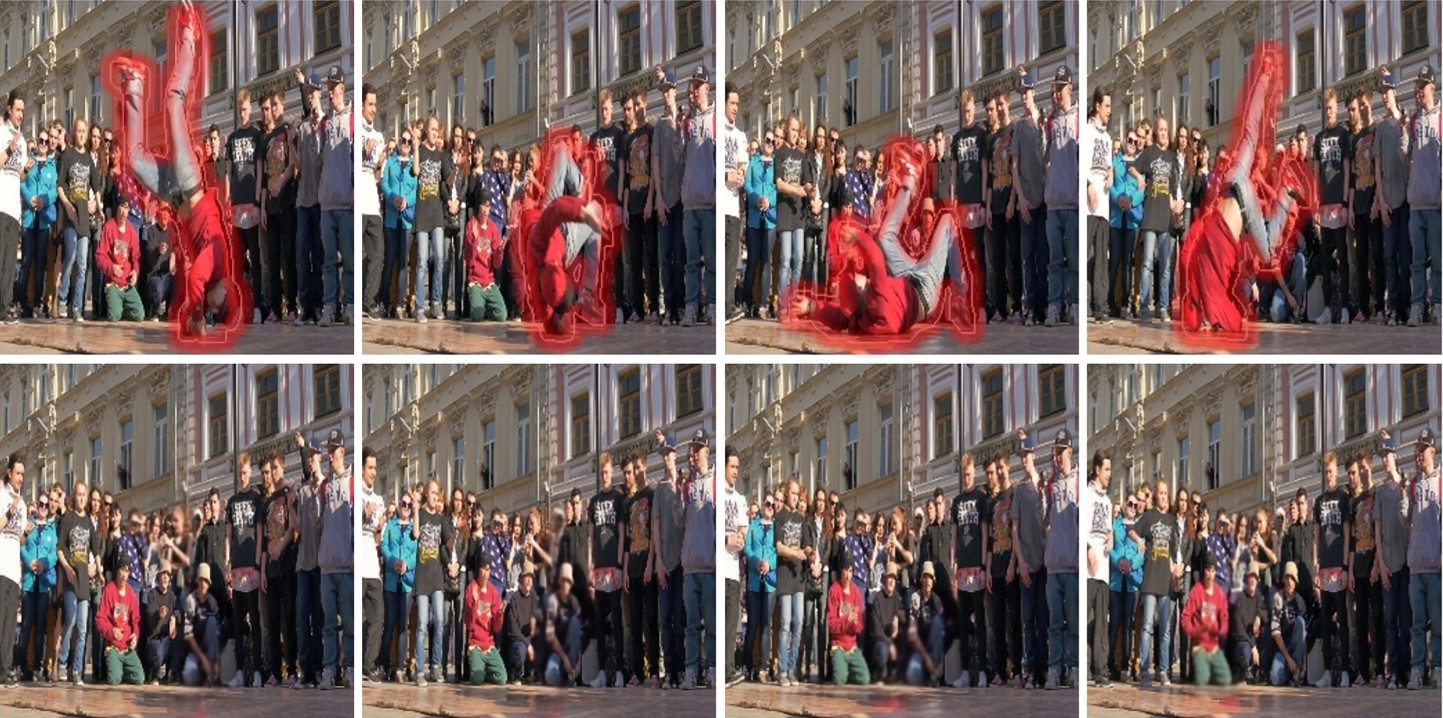}\\
    \includegraphics[width=0.495\linewidth]{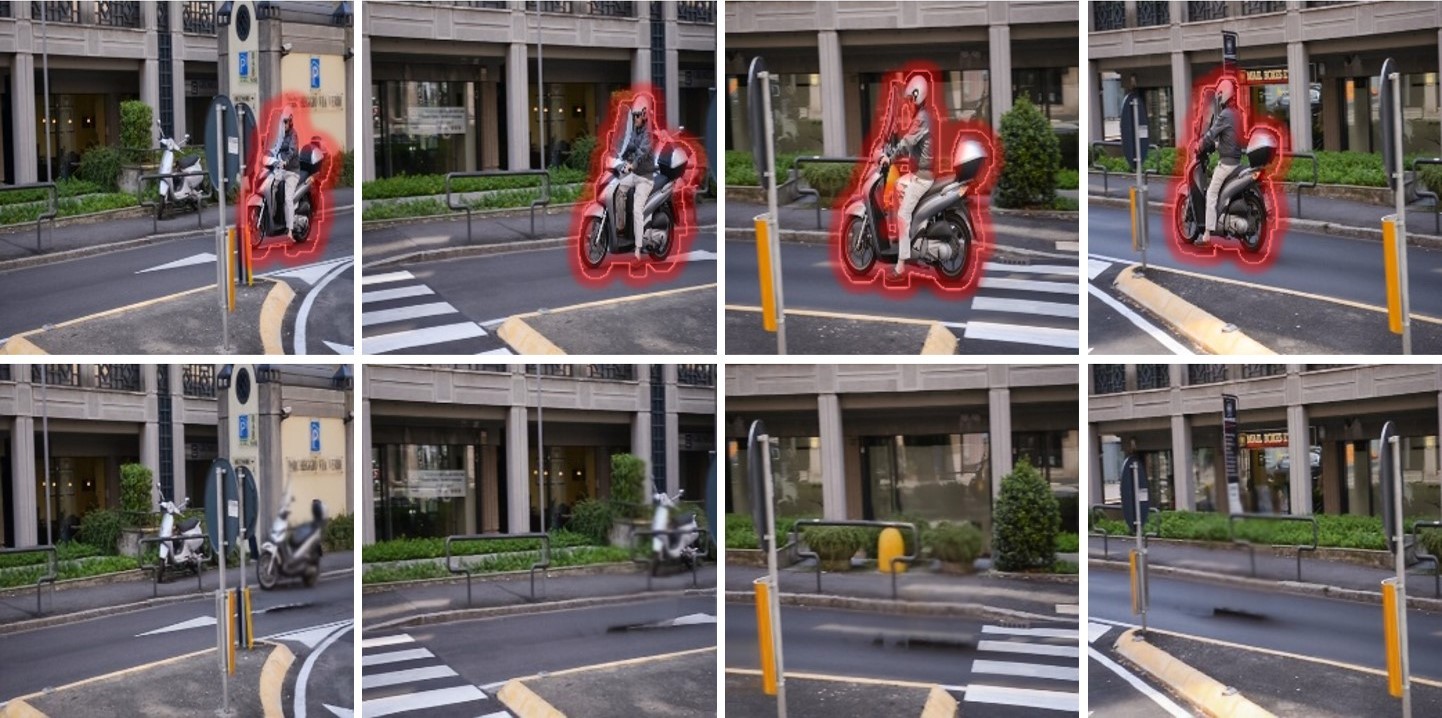} & 		
    \includegraphics[width=0.495\linewidth]{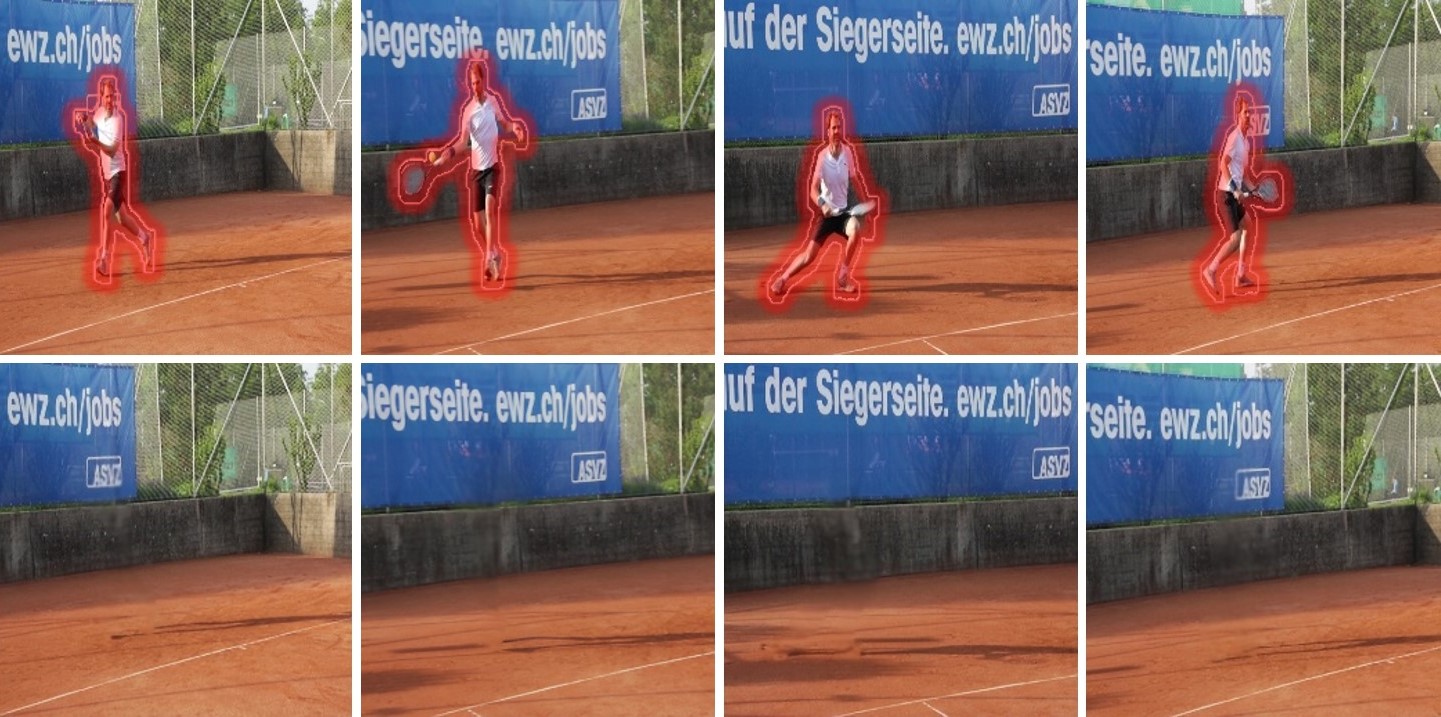}\\
\end{tabular}
\end{center}
\caption{\textbf{Object removal from DAVIS video sequences.} For each input sequence, we show representative frames with mask boundaries in red. We show the inpainted results using our method in even rows.}
\label{fig:obj_removal}
\end{figure*}

\subsection{Spatio-Temporal Video Quality}

Wang~\etal~\cite{wang2018video} proposed a video version of the inception score (FID) to quantitatively evaluate the quality of video generation. We take this metric to evaluate the quality of video inpainting as it measures the spatio-temporal quality in a perceptual level. As in~\cite{wang2018video}, we follow the protocol that uses the I3D network~\cite{carreira2017quo} pretrained on a video recognition task to measure the distance between the spatio-temporal features extracted from the output videos and the ground-truth videos. 
% The FID score between the output videos and the ground-truth is calculated as
%{\small
% \begin{equation}
%     FID = \left \| \mu - \hat\mu \right\|_{2} + \mathrm{Tr}\left( \sum + \hat\sum - 2\sqrt{\sum\hat\sum} \right),
% \end{equation}
% %}
% where $\mu$ and $\sum$ denote the mean and covariance matrix of the extracted feature vectors respectively.

For this experiment, we take 20 videos in the DAVIS dataset. For each video, we ensure to choose a different video out of the other 19 videos to make a mask sequence, so that we have the setting where our algorithm is supposed to recover the original videos rather than remove any parts. We use the first 64 frames for both input and mask videos. 
We run five trials as in~\secref{sec:TC} and average the FID scores over the videos and trials.
\tabref{tab:FID} summarizes the results. Our method has the smallest FID among the compared methods. This implies that our method achieves both better visual quality and temporal consistency.

\begin{figure*}[t]
\begin{center}
\def\arraystretch{1.0}
\begin{tabular}{@{}c@{\hskip 0.008\linewidth}c@{\hskip 0.008\linewidth}c@{}}
    \includegraphics[width=0.193\linewidth]{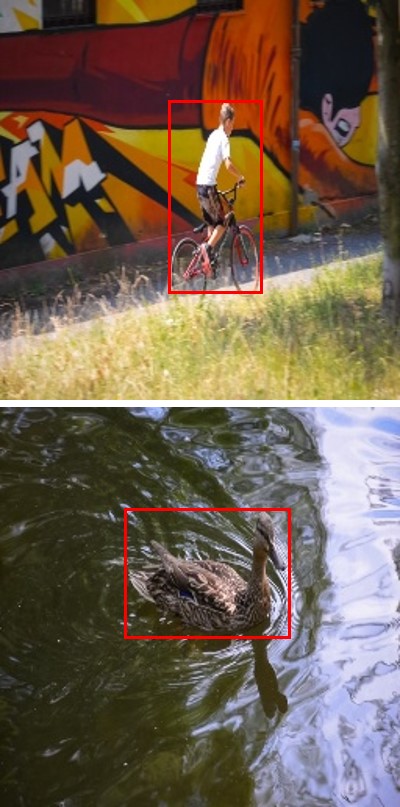}&
    \includegraphics[width=0.397\linewidth]{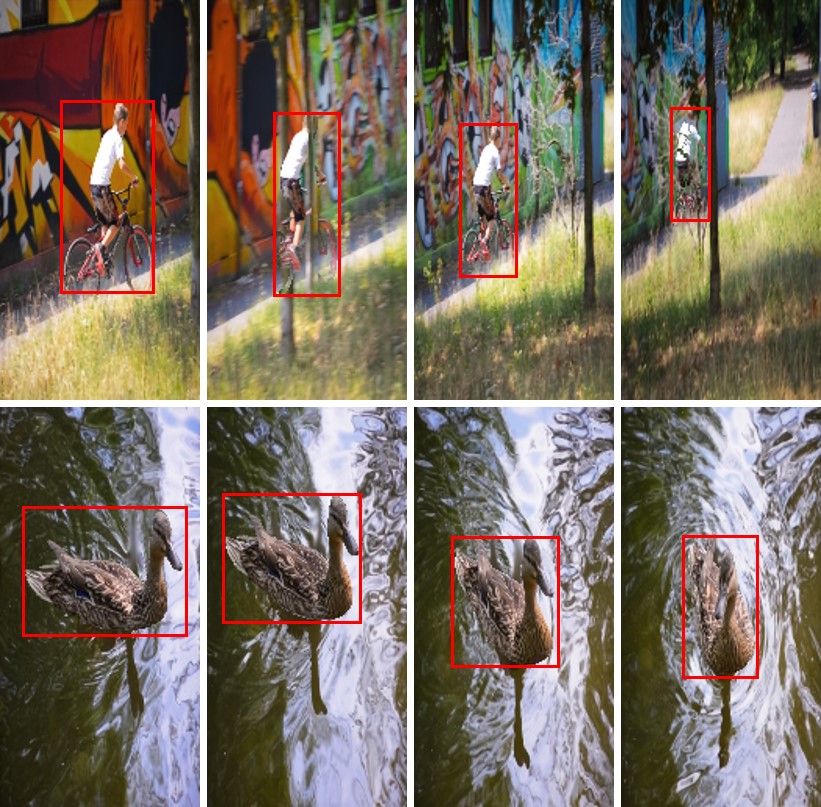}&
    \includegraphics[width=0.39\linewidth]{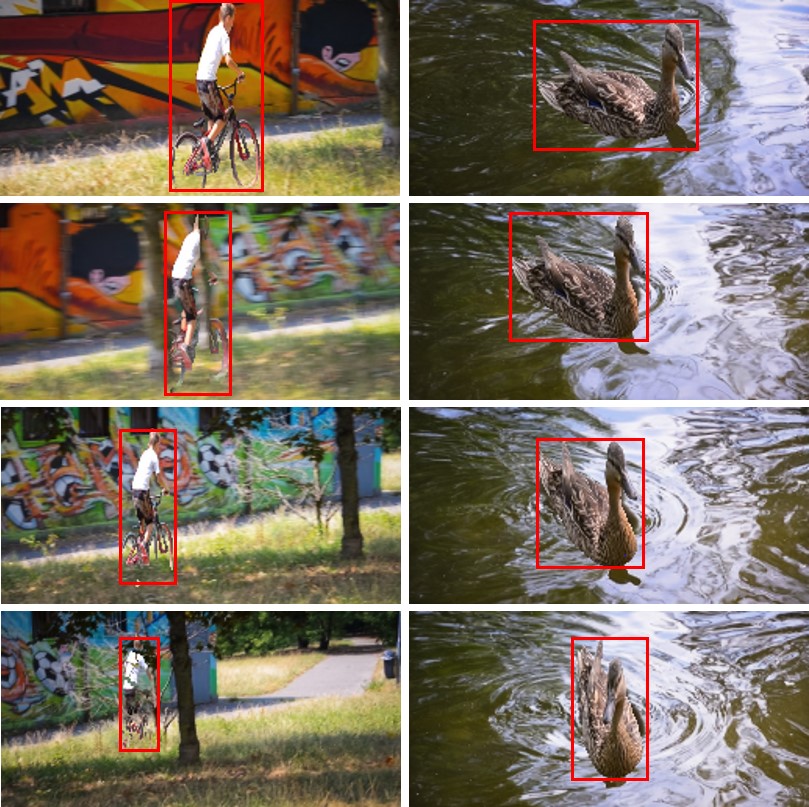}\\
{(a) First input frame} & {(b) Horizontally shrunk frames} & {(c) Vertically shrunk frames} \\
\end{tabular}
\end{center}
%\caption{\textbf{Extension to video retargeting (a-c) and video super-resolution (d-h).} (a) Original first frame. (b) Horizontally shrunk frames. (c) Vertically shrunk frames. (d-e) Inputs of the super-resolution task. (f) Bicubic result. (g) Our SR result (h) ground truth.}
\caption{\textbf{Extension to video retargeting.} (a) Original first frame. (b) Horizontally shrunk frames. (c) Vertically shrunk frames.}
\label{fig:retarget}
\end{figure*}

\begin{figure}[t]
\begin{center}
\def\arraystretch{1.0}
\begin{tabular}{@{}c@{}}
    \includegraphics[width=\linewidth]{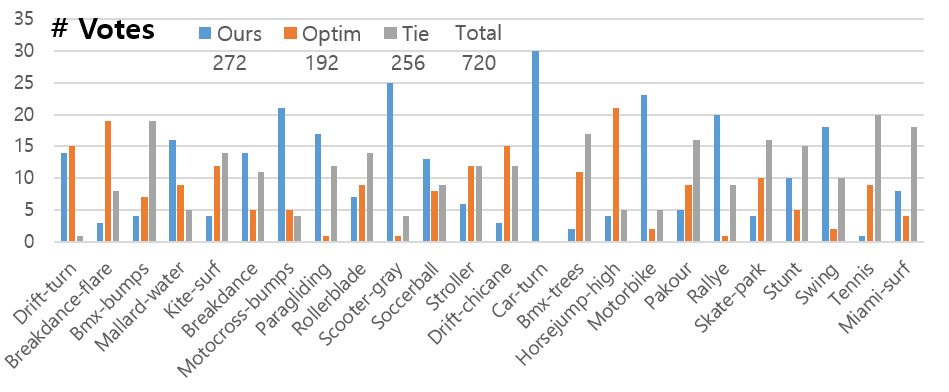}\\
\end{tabular}
\end{center}
\vspace{-6mm}
\caption{\textbf{User study results}.}
\label{fig:user}
\end{figure}

\subsection{User Study on Video Object Removal}
We apply our approach to remove dynamically moving objects in videos. We use 24 videos from the DAVIS dataset~\cite{perazzi2016benchmark,pont20172017} of which the names are listed in \figref{fig:user}. Examples of our results are in \figref{fig:obj_removal}. We perform a human subjective test for evaluating the visual quality of inpainted videos. We compare our method with the strong optimization baseline~\cite{huang2016temporally} which is specifically aimed for the video completion task.

In each testing case, we show the original input video, our removal result and the result of Huang~\etal on the same screen. The order of the two removal video results is shuffled. To ensure that a user has enough time to distinguish the difference and make a careful judge, we play all the video results once at the original speed and then once at $0.5\times$ speed. Also, a user allows seeing videos multiple times. Each participant is asked to choose a preferred result or tie. A total of 30 users participated in this study. We specifically ask each participant to check for both image quality and temporal consistency. The user study results are summarized in \figref{fig:user}. It shows that, while there are different preferences across video samples, our method is preferred more often by the participants.
% We apply our approach to remove dynamically moving objects in videos. We use 24 videos from the DAVIS dataset~\cite{perazzi2016benchmark,pont20172017} of which the names are listed in \figref{fig:user}. Examples of our results are in \figref{fig:obj_removal}. We perform a human subjective test for evaluating the visual quality of inpainted videos. We compare our method with the strong optimization baseline~\cite{huang2016temporally} which is specifically aimed for the video completion task. In each testing case, we show the original input video, our removal result and the result of Huang~\etal on the same screen. The order of the two removal video results is shuffled. To ensure that a user has enough time to distinguish the difference and make a careful judge, we play all the video results once at the original speed and then once at $0.5\times$ speed. Also, a user allows seeing videos multiple times. Each participant is asked to choose a preferred result or tie. A total of 30 users participated in this study. We specifically ask each participant to check for both image quality and temporal consistency. The user study results are summarized in \figref{fig:user}. It shows that, while there are different preferences across video samples, our method is preferred more often by the participants. The complete video results can be found in the supplementary materials.

\subsection{Application to Video Retargeting}
%\noindent \textbf{Video retargeting.} \quad
Video retargeting aims to adjust the aspect ratio (or size) of frames to fit the target aspect ratio while maintaining salient content in a video. We propose to solve video retargeting by \textit{removing and then adding}, which is a potential pipeline where our framework would run in combination with other AR (\ie overlaying) technologies. Specifically, we first remove the salient content by inpainting the background, resize the inpainted frames into the target aspect ratio, and then overlay the salient content after the desirable rescaling. To simplify the settings, we target to horizontally or vertically shrink the frames while keeping the original aspect ratio of the moving object. The saliency masks can be automatically estimated, for example, by a feed-forward CNN~\cite{cho2017weakly}, however we assume a more constrained scenario where the saliency masks are given as the object segmentation masks for all frames. Our method yields little warble and jittering over time and produces natural video sequences. \figref{fig:retarget} shows examples of the retargeted frames.

\subsection{Limitation}
We observe color saturation artifacts when there is a large and long occlusion in a video. The discrepancy error of the synthesized color propagates over time, causing inaccurate warping. The regions that have not been revealed in the temporal radius is synthesized blurry. Also, due to the limited memory footprint, we only experimented with $256 \times 256$ px frames.

\section{Conclusion}
In this paper, we propose a novel framework for video inpainting. Based on the multi-to-single encoder-decoder network, our model learns to aggregate and align the feature maps from neighbor frames to inpaint videos. We use the recurrent feedback and the temporal memory to encourage temporally coherent output. Our extensive experiments demonstrate that our method achieves superior visual quality than the state-of-the-art image inpainting solution and performs favorably against an optimization method both qualitatively and quantitatively. Despite some limitations, we argue that a well-posed feed-forward network has a great potential to avoid computation-heavy optimization method and boosts its applicability in many related vision tasks.

% , while overwhelming the inference speed. +LIMITATION/FUTURE?
% Compared with per-frame application of a state-of-the-art image inpainting algorithm, our video results are of significantly better visual quality. 

\paragraph{Acknowledgements}
Dahun Kim was partially supported by Global Ph.D. Fellowship Program through the National Research Foundation of Korea (NRF) funded by the Ministry of Education (NRF-2018H1A2A1062075).

{\small
\bibliographystyle{ieee}
\bibliography{egbib}
}

\end{document}